\documentclass[letterpaper, 10 pt, conference]{ieeeconf}  

\IEEEoverridecommandlockouts                              

\usepackage{graphics} 
\usepackage{amsmath} 
\usepackage{amssymb}  
\usepackage{booktabs}
\usepackage[table]{xcolor}
\usepackage{xurl}
\usepackage[pdfmajorversion=5]{graphicx}
\newcommand{\rankfirst}[1]{\cellcolor{gray!15}\textbf{#1}}
\newcommand{\ranksecond}[1]{\cellcolor{gray!15}\underline{#1}}
\newcommand{\rankthird}[1]{\cellcolor{gray!15}#1}

\title{\LARGE \bf
RiskWorld: Risk-Aware World Modeling with Flow-Guided Occupancy Evolution for Selective Trajectory Planning in Automated Driving
}

\author{Rongxiang Zeng$^{1,*}$, Linsen Cai$^{1,*}$, Jiafu Zhang$^{1}$, Yijie Zhong$^{1}$, Yide Tao$^{2}$, Shuai Wang$^{3}$, Nan Zheng$^{2}$,\\
Hai L. Vu$^{2}$, Alvaro García Hernandez$^{1}$ and Yongqi Dong$^{1,\dagger}$
\thanks{This work was funded under the Excellence Strategy of the Federal Government and the Länder and with support from the Exploratory Research Space (ERS) of RWTH.}
\thanks{$^{1}$Rongxiang Zeng, Linsen Cai, Jiafu Zhang, Yijie Zhong, Alvaro García Hernandez and Yongqi Dong are with
        RWTH Aachen University, Aachen, Germany.
        }
\thanks{$^{2}$Yide Tao, Nan Zheng and Hai L. Vu are with
        Monash University, Melbourne, Australia.}%
\thanks{$^{3}$Shuai Wang is with
        the Technical University of Munich (TUM), Munich, Germany.}%
\thanks{$^{*}$Rongxiang Zeng and Linsen Cai contributed equally to this work.}%
\thanks{%
$^{\dagger}$\,Corresponding author: Yongqi Dong%
\protect\\
\phantom{$^{\dagger}$\,}\hspace{1.11em}E-mail: \texttt{yongqi.dong@rwth-aachen.de}.%
}
}

\begin{document}

\maketitle
\thispagestyle{empty}
\pagestyle{empty}

\begin{abstract}

Safe motion planning in automated driving requires anticipating evolving traffic risks and
deciding when to revise the current planned trajectory. We introduce RiskWorld,
a risk-aware world modeling framework for shared occupancy forecasting and selective
trajectory replacement. Spatial risk fields and temporal actor context
are fused with visual bird's-eye-view features. Flow-guided evolution
transports occupancy and scene features, while signed residuals correct occupancy after transport. 
One forecast is generated per planning step and reused across candidates.
Each candidate is compared with a current-state persistence reference, yielding a nonnegative collision-score correction. 
The trajectory selected by current-world evaluation serves as the planning anchor and is replaced only when additional predicted risk triggers intervention and an alternative satisfies component-wise constraints on predicted risk and trajectory error. Candidate geometries remain unchanged. We evaluate RiskWorld for open-loop planning on nuScenes using camera features, annotation-derived current and historical actor states, and dataset-provided map context. 
RiskWorld achieves the lowest collision rate at a long evaluation horizon of 3 s, and the second-best average L2 error among various state-of-the-art baselines, while running at 11.5 FPS on a single NVIDIA RTX 4090 with 90.81\,M parameters.
Within-setting ablations show that RiskWorld achieves lower collision
rates than the current-state rescoring baseline, while forecast reuse
enables additional candidates to be evaluated at low marginal
computational cost.

\end{abstract}

\section{Introduction}
\label{sec:introduction}
As automated vehicles (AVs) are increasingly deployed on public roads,
ensuring safe motion planning of AVs has become a critical requirement.
Road safety broadly concerns both reducing the likelihood of collisions
and mitigating the severity of their consequences.
For motion planning of AVs, this requires anticipating how surrounding road users
may evolve over the planning horizon.
A trajectory that is collision-free regarding the current scene may
still intersect another actor's future path~\cite{cui2021lookout,pini2023safepathnet}. For example, a vehicle in an
adjacent lane may move into the ego vehicle's intended path during execution.
Motion planning should therefore account for the temporal evolution of the
traffic scene rather than assess collision risk solely from the current
observation.

Bird's-eye-view (BEV) representations place road geometry and surrounding
actors in a common spatial frame, supporting the integration of perception,
prediction, and ego-trajectory planning~\cite{hu2022stp3,hu2023uniad}.
Driving risk field (DRF)-based models~\cite{10394462} complement
geometric scene representations by encoding the spatial distribution
of risk associated with road geometry and the positions and motion
of surrounding road users. These representations provide a basis for evaluating
spatial relationships and motion-related exposure. Occupancy
forecasting~\cite{hu2021fiery,mahjourian2022occupancy} and world
modeling~\cite{zheng2024occworld,zhangResWorldTemporalResidual2026} extend
this reasoning over time, providing future scene states that can inform
trajectory evaluation.

Predicting a plausible future does not by itself justify changing a
selected trajectory. A future-risk score may reflect hazards already
captured by the current-scene evaluation or additional exposure caused by
scene evolution. Planning must also account for uncertainty in predicted
actor motion~\cite{cui2021lookout}. These considerations motivate comparing
future evidence with a current-state reference before deciding whether to
intervene.

We propose \textbf{RiskWorld}, a risk-aware world modeling method for shared
occupancy forecasting and selective trajectory replacement. Visual BEV
features and structured current and historical risk initialize a flow-guided
rollout that transports occupancy and scene features, with signed residual
corrections for changes not explained by displacement. A single forecast
is computed per planning step and shared across candidates, avoiding
repeated world evolution. Each candidate queries the forecast and a persistence reference
that repeats the current state at corresponding locations. Their contrast
yields a nonnegative correction to its current-world collision score. The
trajectory selected by the current-world evaluation serves as the anchor
and is retained unless additional predicted risk triggers intervention
and an alternative satisfies component-wise constraints on predicted risk
and trajectory error. Candidate geometries remain
unchanged.

We evaluate RiskWorld on nuScenes~\cite{caesar2020nuscenes} with
camera features, annotation-derived current and historical actor
states, and dataset-provided map context supplied at inference.
Within-setting studies examine future-risk intervention and spatial
transport, with additional analyses of temporal order in incoming-occupancy prediction
and the computational cost of forecast reuse.

In short, we make three main contributions:
\begin{itemize}
\item We introduce a \textbf{risk-aware world representation} that aligns
visual BEV features with spatial hazard fields and temporal actor context
for future-scene prediction and trajectory-level risk assessment.

\item We develop a \textbf{shared flow-guided occupancy model} that
jointly transports occupancy and scene features, with signed residual
corrections for occupancy changes not explained by transport.

\item We propose a \textbf{persistence-referenced intervention rule} that
converts future--current evidence contrasts into nonnegative collision-score
corrections for selective trajectory replacement within a fixed candidate
set, subject to constraints on predicted risk and trajectory error.
\end{itemize}

\section{Related Works}
\label{sec:related_works}
\label{sec:related_work}

\subsection{Planning-Oriented Representations and Risk Modeling}
Planning-oriented representations organize scene information around the
requirements of ego-motion planning. ST-P3~\cite{hu2022stp3},
UniAD~\cite{hu2023uniad}, and VAD~\cite{jiang2023vad} couple perception
and motion prediction with trajectory planning, while
SparseDrive~\cite{sunSparseDriveEndtoEndAutonomous2025} uses sparse
instances to integrate detection, tracking, online mapping, and planning.
Geometric BEV encoders such as GeoBEV~\cite{zhang2025geobev} provide
spatial features for scene understanding. Although their representations
and task organization differ, these approaches share the aim of making
scene structure and actor motion useful to the planner.

Risk-field methods provide a complementary account of spatial exposure.
DRF-based planning~\cite{10394462} represents hazards associated with
road geometry and traffic-participant states, whereas
Kolekar et al.~\cite{kolekar2020risk} relate driver-perceived risk to
corrective actions. Geometric and risk-based representations thus serve
related but distinct purposes: the former describe the spatial arrangement
of the scene, while the latter express its implications for driving risk.
Extending this assessment over the planning horizon requires considering
how actor motion changes exposure along candidate trajectories.

\subsection{Occupancy and World Forecasting}
Occupancy forecasting provides a spatial representation of future
traffic. FIERY~\cite{hu2021fiery} predicts future BEV instances, while
Occupancy Flow Fields~\cite{mahjourian2022occupancy} and
ImplicitO~\cite{agro2023implicit} couple occupancy and motion for
spatiotemporal reasoning. Building on occupancy representations,
OccWorld~\cite{zheng2024occworld},
RenderWorld~\cite{yan2025renderworld}, and
OccLLaMA~\cite{wei2024occllama} connect scene forecasting with
ego-motion prediction, whereas Drive-WM~\cite{wang2024drivewm}
supports planning through multi-view visual forecasting.

A complementary modeling approach expresses scene evolution through
spatial transport. EfficientOCF~\cite{xu2025efficientocf} uses predicted
flow to associate dynamic instances across time, while
DFIT-OccWorld~\cite{zhang2024dfitoccworld} warps current occupancy
features and refines the resulting representation to predict future
occupancy. The value of these forecasts for planning also depends on
how predicted scene information is incorporated into trajectory
evaluation or revision.

\subsection{Prediction-to-Planning Interfaces}
Forecasts inform planning through trajectory evaluation, joint prediction
and planning, or trajectory decoding. LookOut~\cite{cui2021lookout} and
SafePathNet~\cite{pini2023safepathnet} incorporate predicted traffic
motion into trajectory evaluation and selection.
ScePT~\cite{chen2022scept} and DTPP~\cite{huang2024dtpp} condition actor predictions on ego motion,
allowing candidate-dependent interactions to inform planning. This
conditioning represents potential responses to ego actions, whereas a
shared forecast evaluates candidates against the same predicted
environment.

Future representations can also guide trajectory generation or refinement.
World4Drive~\cite{zheng2025world4drive} uses latent futures for trajectory
generation, while ResWorld~\cite{zhangResWorldTemporalResidual2026} uses
temporal residuals for future-guided refinement.
IR-WM~\cite{mei2026irwm} examines alternative forecasting--planning
couplings. A complementary approach is to check an existing plan before
invoking an alternative. SafetyNet~\cite{vitelli2022safetynet} follows
this approach by checking a learned policy and intervening when required.

A remaining challenge in forecast-based planning is to distinguish
additional predicted exposure from risk already represented in the
current-scene assessment and determine whether it warrants changing
the selected trajectory. The proposed \textbf{RiskWorld} addresses this challenge
by combining risk-aware BEV features with a shared occupancy forecast
and comparing candidate-specific future evidence with a current-state
persistence reference. The resulting nonnegative collision-score
corrections guide selective replacement within a fixed candidate set,
subject to constraints on predicted risk and predicted trajectory error.

\section{Method}
\label{sec:method}
\subsection{Problem Setup and Framework Overview}
\label{sec:preliminaries}

\begin{figure*}[t]
    \centering
    \setlength{\abovecaptionskip}{2pt}
    \setlength{\belowcaptionskip}{-4pt}
    \includegraphics[
        width=0.99\textwidth,
        trim=0 1 0 1,
        clip
    ]{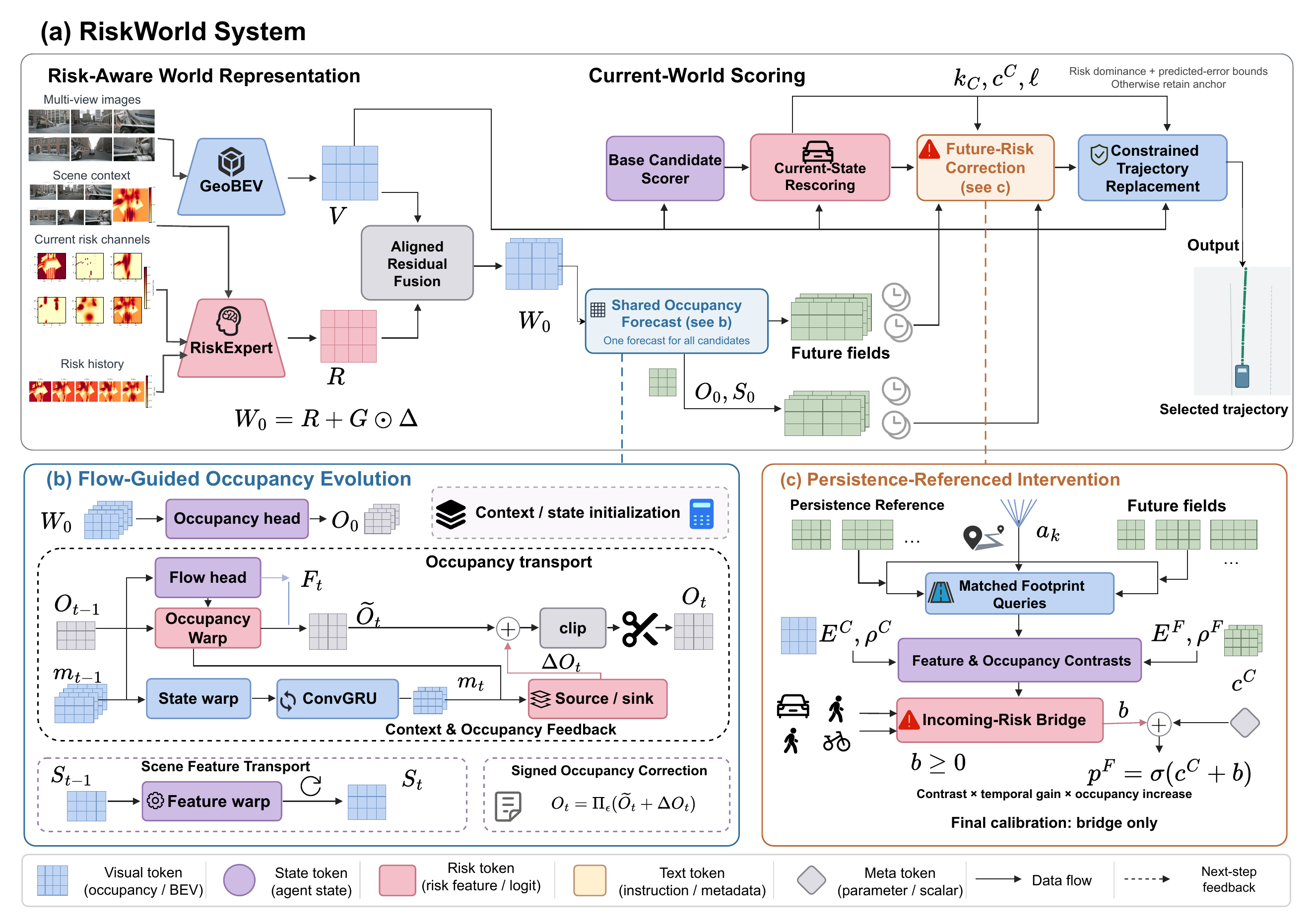}
    \vspace{-3mm}
    \caption{\footnotesize
    Overview of RiskWorld.
    (a) RiskExpert fuses risk history, scene context, and visual BEV features
    for candidate evaluation and shared forecasting.
    (b) A shared displacement field transports occupancy, motion, and scene
    features, followed by signed occupancy correction.
    (c) Candidates query the forecast and current-state persistence at matched
    locations; their contrasts yield nonnegative collision-score corrections
    for constrained trajectory replacement. Only the incoming-risk bridge is
    optimized during final calibration.
    }
    \label{fig:framework}
    \vspace{-4mm}
\end{figure*}

Fig.~\ref{fig:framework} provides an overview of RiskWorld, with
Fig.~\ref{fig:framework}(a) showing the processing pipeline. At each planning step,
RiskWorld selects an ego trajectory from a finite candidate set
based on current-state information and a shared forecast of scene
evolution. The inputs comprise visual BEV features, current and
historical risk fields, map context, actor and ego states, and a
navigation command. Structured actor and map information is supplied
separately from the visual features.


To represent the driving risk field spatially, we define a nonnegative scalar potential over the BEV domain:
\begin{equation}
    r_t:\Omega\rightarrow\mathbb{R}_{\geq0},
    \qquad
    \mathbf{x}\mapsto r_t(\mathbf{x}),
    \label{eq}
\end{equation}
where $\Omega\subset\mathbb{R}^{2}$ denotes the spatial BEV domain, and $\mathbf{x}=(x,y)\in\Omega$ denotes the specific spatial location.
In RiskWorld, road geometry, actor occupancy, and motion-related risk potentials are rasterized into separate BEV channels. These channels encode scene structure and spatial risk cues rather than calibrated collision probabilities.

The candidate set is constructed using a trajectory codebook learned
from training-set ground-truth ego trajectories by clustering residuals
relative to a straight-line reference that maintains current ego
speed. At inference, the corresponding group is selected based
on ego speed and navigation command. Its ten residual prototypes
are added to the reference trajectory and augmented with two deterministic
lower-speed kinematic candidates, to form a set $K=12$ candidates:
\begin{equation}
    \mathcal{A}=\{a_k\}_{k=1}^{K},
    \qquad
    a_k=\bigl((x_{k,t},y_{k,t},\theta_{k,t})\bigr)_{t=1}^{T}.
    \label{eq:candidate_definition}
\end{equation}
Here, $(x_{k,t},y_{k,t})$ and $\theta_{k,t}$ denote the planned ego
position and heading, respectively, for candidate $k$ at future step $t$,
expressed in the current ego frame. All candidates span a $3\,\mathrm{s}$
horizon with $T=6$ steps at $0.5\,\mathrm{s}$ intervals, matching the
temporal resolution of the future supervision and enabling risk evaluation
over the $1$, $2$, and $3\,\mathrm{s}$ prefixes.

To evaluate these candidates, RiskWorld constructs a shared world
representation from the risk and visual inputs. RiskExpert encodes the risk fields with map and motion context. Its
features are aligned and fused with visual BEV features to form the
current world representation $W_0$.
A \emph{base candidate scorer} (BCS) predicts collision scores and
trajectory errors; \emph{current-state rescoring} (CSR) updates this
evaluation using current-state evidence and selects an anchor trajectory.
In parallel, $W_0$ initializes the occupancy rollout
in Fig.~\ref{fig:framework}(b).
The flow head predicts a displacement field $F_t$
from the previous motion state $m_{t-1}$.
This field drives the warping of occupancy, motion
state, and scene features, followed by a signed
occupancy correction.
The resulting occupancy and scene features
$(O_t,S_t)$ form the future fields queried by
the candidates.

The rollout is computed once per planning step and reused across
candidates. Candidates determine query locations, not actor responses,
in contrast to action-conditioned
prediction~\cite{yang2025driveoccworld,du2026sparseworldtc}.
\emph{Persistence-referenced intervention} (PRI) compares each candidate's
forecast queries with queries of a repeated current-state reference at
identical locations (Fig.~\ref{fig:framework}(c)). Feature and occupancy
contrasts yield nonnegative collision-score corrections. PRI retains
the anchor unless additional predicted risk triggers intervention and
an alternative satisfies component-wise constraints on predicted risk and trajectory error. Selection leaves candidate
geometries unchanged.

\subsection{Risk-Aware World Representation}
\label{sec:risk_representation}

\textbf{Risk-field construction:}
At each observed time $\tau$, actor states and semantic map layers are
rasterized in the current reference sensor frame to form a multi-channel
risk tensor $\mathcal{R}_{\tau}\in\mathbb{R}^{C_r\times H_r\times W_r}$
where $C_r$ denotes the number of configured risk channels, and
$H_r$ and $W_r$ denote the height and width of the BEV risk grid,
respectively. The channels encode static-map risk, actor occupancy,
motion-induced risk, and pairwise interaction risk. Static-map risk
combines non-drivable-area penalties with proximity potentials associated
with road/lane dividers, pedestrian crossings, and road edges. Actor
occupancy is obtained by rasterizing oriented actor footprints, taking
their union within the vehicle, pedestrian, and cyclist groups, and
summing the resulting group masks.

For actor $i$, let $u_i$ and $v_i$ denote the longitudinal and lateral
offsets of location $\mathbf{x}$ from the actor center, expressed in its
local heading frame. Its motion-induced risk potential is defined as
\begin{equation}
\phi_i(\mathbf{x})=A_i\,\omega_i(\mathbf{x})
\exp\!\left[
-\frac{u_i^2}{s_{i,x}^2+\epsilon_{\mathrm{ker}}}
-\frac{v_i^2}{s_{i,y}^2+\epsilon_{\mathrm{ker}}}
\right],
\label{eq:drf_motion_kernel}
\end{equation}
where $A_i$ is a class-dependent amplitude, $\omega_i(\mathbf{x})$ is a
speed-dependent directional weighting term, and $s_{i,x}$ and
$s_{i,y}$ denote the longitudinal and lateral spatial extents determined
by actor dimensions, motion state, and class. The constant
$\epsilon_{\mathrm{ker}}>0$ ensures numerical stability. After spatial
truncation and drivable-area weighting, the actor-wise potentials
$\phi_i$ are accumulated by actor class and normalized to form the
vehicle- and vulnerable-road-user (VRU)-motion channels of
$\mathcal{R}_{\tau}$. 
Here, VRUs comprise pedestrians and bicycles.

Pairwise-interaction and TTC-inspired fields use constant-velocity closest
approach, placing spatial potentials between extrapolated actor positions
and weighting them by predicted clearance, time, and actor class.

An aggregate field combines the raw static, occupancy, motion, and
interaction potentials before normalization. Composite fields are
normalized by $\mathcal{N}$ using logarithmic compression, 5th--95th
spatial-percentile scaling, and clipping to $[0,1]$. A degenerate
percentile range falls back to maximum-based scaling, while zero fields
remain unchanged. Channel selection and field parameters follow the
preprocessing configuration.

\textbf{Spatiotemporal encoding and fusion:}
A shared convolutional encoder maps each risk frame to a spatial token grid.
Temporal self-attention aggregates history at each grid location, while a residual
multilayer perceptron incorporates map tokens. Separate gated recurrent
units encode actor and ego histories, which enter through cross-attention
with risk tokens serving as queries. A spatial residual mixer then combines information from neighboring tokens.

Visual features are obtained from GeoBEV~\cite{zhang2025geobev}.
Risk features are resampled onto the visual grid using the frame transform,
and projections align their channel dimensions. Let $R,V$ denote the aligned
risk and visual features, and let $M$ mark risk-grid coverage. The fusion module predicts a feature residual $\Delta$ that updates
the risk representation using the aligned visual and risk features:
\begin{equation}
    \Delta=f_{\Delta}\!\left([V,R,|V-R|,V\odot R,M]\right),
    \label{eq:visual_risk_residual}
\end{equation}
where brackets denote channel concatenation and $\odot$ is element-wise
multiplication. It is added to the risk representation with a coverage-dependent weight to obtain the current world representation:
\begin{equation}
    W_0=R+G\odot\Delta,
    \qquad G=gM+(1-M),
    \label{eq:risk_visual_fusion}
\end{equation}
with $G$ broadcast across channels. 

The fixed coefficient $g\in(0,1)$
scales the residual within risk coverage, while the full residual is applied
elsewhere. The resulting representation $W_0$ supports both current-world
candidate scoring and occupancy evolution.

\subsection{Flow-Guided Occupancy Evolution}
\label{sec:occupancy_evolution}

We model scene evolution through spatial transport
and occupancy correction, drawing on flow-based temporal
association~\cite{xu2025efficientocf} and residual
state prediction~\cite{mei2026irwm}.
Transport propagates existing scene structure,
while residual correction accounts for
occupancy changes not captured by warping alone. To initialize the rollout, the fused representation $W_0$ is mapped to scene features
$S_0=f_{\mathrm{scene}}(W_0)$, from which current occupancy is predicted as current occupancy
$O_0=\sigma(f_O(S_0))$. The motion state is initialized
as $m_0=C=S_0+f_C(W_0)$, with $C$ held fixed throughout the rollout
to provide current-scene context alongside the evolving state.

At step $t$, the flow head predicts
$F_t=s_F\tanh(f_F(m_{t-1}))$,
where $s_F$ bounds displacement in meters.
The backward warp $\mathcal{W}(X,F)$ bilinearly
samples $X$ at $\mathbf{x}-F(\mathbf{x})$.
We use the same displacement for occupancy,
motion state, and scene features:
\begin{equation}
    \begin{aligned}
        \widetilde{O}_t
        &=\Pi_{\epsilon}\!\left(
            \mathcal{W}(O_{t-1},F_t)\right),\\
        \widetilde{m}_t
        &=\mathcal{W}(m_{t-1},F_t),\\
        S_t
        &=\mathcal{W}(S_{t-1},F_t),
    \end{aligned}
    \label{eq:shared_transport}
\end{equation}
where $\Pi_{\epsilon}$ clips predicted occupancy values to
$[\epsilon,1-\epsilon]$.
Sharing the warp gives the propagated quantities
a common spatial correspondence, so subsequent
trajectory queries sample occupancy and features from the same source locations.

The motion update combines the transported state
with the initial context and occupancy feedback:
\begin{equation}
    m_t=\mathcal{U}\!\left(
        C+e_t+f_{\mathrm{fb}}(\widetilde{O}_t-O_0),
        \widetilde{m}_t
    \right),
    \label{eq:motion_recurrence}
\end{equation}
where $e_t$ is a spatially broadcast time embedding,
$f_{\mathrm{fb}}$ projects the occupancy difference, and
$\mathcal{U}$ is implemented with a ConvGRU.
The feedback
$\widetilde{O}_t-O_0$ captures the departure of transported occupancy
from the current prediction, allowing recurrent updates to incorporate
occupancy evolution alongside the initial scene context.

Spatial transport alone does not account for
all changes in occupied space.
Two heads then predict $B_t=f_{\mathrm{src}}(m_t)$ and
$D_t=f_{\mathrm{snk}}(m_t)$, yielding the signed
occupancy correction
$\Delta O_t=\alpha[\tanh(B_t)-\tanh(D_t)]$.
Occupancy is updated as
\begin{equation}
    O_t=\Pi_{\epsilon}\!\left(
        \widetilde O_t+\Delta O_t
    \right),
    \label{eq:occupancy_correction}
\end{equation}
where $\alpha$ controls the correction magnitude. The bounded
correction allows occupancy to increase or decrease after transport
without modifying the transported scene features $S_t$.
Repeating these updates yields the shared future sequence
$\{(O_t,S_t)\}_{t=1}^{T}$ for candidate evaluation.

\subsection{Persistence-Referenced Intervention} 
\label{sec:conservative_planning}
The shared future sequence $\{(O_t,S_t)\}_{t=1}^{T}$ is used to
assess whether to replace the CSR-selected anchor $a_{k_C}$.
For each candidate $k$, CSR provides the current-state collision logits
$c^C_{k,t}$ and predicted prefix trajectory errors
$\boldsymbol{\ell}_k$. Reference ego trajectories define the error
targets during training, whereas inference uses only predicted errors. We consider incoming-risk correction only, with outgoing-risk
release and future-error correction disabled.

\textbf{Future-risk correction:}
Each candidate queries $(O_t,S_t)$ and a persistence reference that repeats
$(O_0,S_0)$ at matched locations. Query patches and footprint masks remain
aligned with the current ego frame. A shared evidence encoder with matched
candidate conditioning produces $E^F_{k,t}$ and $E^C_{k,t}$, while maximum
occupancy over valid footprint cells gives $\rho^F_{k,t}$ and $\rho^C_{k,t}$.
Define $\delta E_{k,t}=E^F_{k,t}-E^C_{k,t}$ and
$\delta\rho_{k,t}=\rho^F_{k,t}-\rho^C_{k,t}$.
The \emph{incoming-risk bridge} uses a temporal encoder to map the
feature contrasts to $z_{k,t}$, yielding the nonnegative gain
$\gamma_{k,t}=[\operatorname{softplus}(z_{k,t})-\log 2]_+$
for the collision-logit correction.
The collision-score update is
\begin{equation}
\begin{aligned}
b_{k,t}
 &=\beta\gamma_{k,t}
 \tanh\!\left(\frac{\|\delta E_{k,t}\|_2}{\sqrt d}\right)
 [\delta\rho_{k,t}]_+,\\
p^F_{k,t}&=\sigma(c^C_{k,t}+b_{k,t}),
\end{aligned}
\label{eq:incoming_risk_calibration}
\end{equation}
where $d$ is the feature dimension, $\beta\geq0$ scales the correction, and
$[x]_+=\max(x,0)$. By construction, $p^F_{k,t}\geq p^C_{k,t}=\sigma(c^C_{k,t})$.
A positive correction requires both nonzero feature contrast and increased
maximum footprint occupancy, not merely a local occupancy
change.

\textbf{Constrained replacement:}
Candidate risk is summarized over prefixes $\mathcal{H}=\{1,2,3\}\,\mathrm{s}$
and the final two steps:
\begin{equation}
\begin{aligned}
r_{k,h}&=\frac{1}{n_h}\sum_{t=1}^{n_h}p_{k,t},
 \qquad h\in\mathcal{H},\\
r_{k,\mathrm{tail}}&=1-(1-p_{k,T-1})(1-p_{k,T}),
\end{aligned}
\label{eq:planning_risk_summary}
\end{equation}
where $n_h$ is the prefix length. The tail term retains late-horizon
evidence and is used as a decision score, not a calibrated joint-event
probability. Applying these summaries to $p^F$ and $p^C$ gives
$\mathbf{r}^F_k$ and $\mathbf{r}^C_k$, including the tail component.

Intervention requires a positive anchor correction and, for some component $j$,
either
$r^F_{k_C,j}>r^C_{k_C,j}+\eta$ or
$r^C_{k_C,j}\leq\tau_j<r^F_{k_C,j}$, where $\eta$ is the margin
and $\tau_j$ are component-specific thresholds. A valid alternative $k\neq k_C$ must satisfy
\begin{equation}
\begin{aligned}
\mathbf{r}^F_k&\leq\mathbf{r}^F_{k_C},\\
\exists j:\quad r^F_{k,j}+\eta&<r^F_{k_C,j},\\
\bar{\ell}_k&\leq\bar{\ell}_{k_C}+\delta_{\mathrm{mean}},\\
\boldsymbol{\ell}_k&\leq\boldsymbol{\ell}_{k_C}
                       +\delta_{\mathrm{prefix}}.
\end{aligned}
\label{eq:replacement_constraints}
\end{equation}
All vector inequalities are component-wise. $\bar{\ell}_k$ is the normalized
weighted mean of predicted prefix errors. Among eligible candidates, we select the one minimizing
\begin{equation}
    J_k = w_{\max}\max_j r^F_{k,j}
        + w_{\mathrm{avg}}\bar r^F_k
        + \bar\ell_k,
    \label{eq:selection_objective}
\end{equation}
where $\bar r^F_k$ is the weighted prefix risk, and
$w_{\max}$ and $w_{\mathrm{avg}}$ weight the risk terms in candidate ranking. Equal prefix weights recover arithmetic means. If intervention is
not triggered or no alternative is eligible, the anchor is retained. The constraints act on predicted quantities and do not
guarantee collision avoidance.

\subsection{Training Objectives and Optimization}
\label{sec:training}

The training of RiskWorld starts from pretraining RiskExpert through future risk-field
reconstruction. Training then proceeds to risk-visual fusion and current-world scoring, followed
by occupancy evolution. Evolution uses occupancy supervision, masked
smooth-L1 displacement loss, and balanced binary cross-entropy (BCE) on
source/sink event targets derived from future annotations.

During final calibration, the representation, evolution module,
evidence encoders, and current-world scorer are frozen. Feature and occupancy contrasts are
detached, and only the incoming-risk bridge is optimized, leaving forecasts and
predicted trajectory errors unchanged. The objective is
\begin{equation}
\begin{aligned}
\mathcal{L}_{\mathrm{cal}}
={}&\mathcal{L}_{\mathrm{BCE}}+\mathcal{L}_{\mathrm{Brier}}
   +\lambda_f\mathcal{L}_{\mathrm{focal}}\\
 &+\mathcal{L}_{\mathrm{hard}}+\lambda_a\mathcal{L}_{\mathrm{aux}}
   +\lambda_r\langle b_{k,t}^{2}\rangle_{\mathrm{valid}}.
\end{aligned}
\label{eq:calibration_objective}
\end{equation}

BCE supervises per-step collision predictions, while the Brier-type
loss penalizes squared prefix-level prediction errors. Focal loss
emphasizes difficult candidate steps, and the hard-example term
focuses on the largest focal losses within each sample. The quadratic
penalty on $b_{k,t}$, averaged over valid candidate-time pairs,
discourages large collision-logit corrections.

The auxiliary objective combines feasibility ranking, false-safe
penalties, prefix and late-horizon hazard supervision for the
BCS-selected candidate, and pairwise comparisons against it.
Annotated trajectory errors restrict training pairs to a reference-error
window and are not used at inference. Invalid candidates and missing
labels are masked, while prefix- and decision-level supervision require
complete future annotations. The incoming-risk bridge is optimized using AdamW with
event-balanced sampling.


\section{Experiments and Results}
\label{sec:experiments}
\begin{table*}[t]
\caption{Planning performance and computational efficiency on nuScenes dataset.}
\label{tab:model_comparison}
\centering
\small
\setlength{\tabcolsep}{2.55pt}
\renewcommand{\arraystretch}{1.12}

\begingroup
\small
\setlength{\tabcolsep}{2.55pt}
\renewcommand{\arraystretch}{1.0}

\begin{tabular}{@{}l|cc|rr|rrrr|rrrr@{}}
\toprule
Method
& Input
& Aux. Sup.
& Params.
& FPS$_{\text{4090}}$
& \multicolumn{4}{c|}{L2 (m) $\downarrow$}
& \multicolumn{4}{c}{CR (\%) $\downarrow$} \\
\cmidrule(lr){6-9}
\cmidrule(lr){10-13}
& & &
(M) $\downarrow$
& $\uparrow$
& 1s & 2s & 3s & Avg.
& 1s & 2s & 3s & Avg. \\
\midrule

ST-P3~\cite{hu2022stp3}
& C & M+B+D
& -- & 1.60
& 1.33 & 2.11 & 2.90 & 2.11
& 0.23 & 0.62 & 1.27 & 0.71 \\

UniAD~\cite{hu2023uniad}
& C & M+B+Mo+Tr+O
& 125.00 & 1.80
& 0.44 & 0.67 & 0.96 & 0.69
& 0.04 & 0.08 & 0.23 & 0.12 \\

VAD~\cite{jiang2023vad}
& C & M+B+Mo
& \rankfirst{58.36} & 4.50
& 0.41 & 0.70 & 1.05 & 0.72
& 0.07 & 0.17 & 0.41 & 0.22 \\

GenAD~\cite{zheng2024genad}
& C & M+B+Mo
& -- & --
& 0.28 & 0.49 & 0.78 & 0.52
& 0.08 & 0.14 & 0.34 & 0.19 \\

SparseDrive-S~\cite{sunSparseDriveEndtoEndAutonomous2025}
& C & M+B+Mo+Tr
& 85.90 & 9.00& 0.29 & 0.58 & 0.96 & 0.61
& \rankfirst{0.01} & \ranksecond{0.05}
& \rankthird{0.18} & \ranksecond{0.08} \\

DiffusionDrive~\cite{liao2025diffusiondrive}
& C & M+B+Mo+Tr
& 92.04 & 8.20
& 0.27 & 0.54 & 0.90 & 0.57
& \rankthird{0.03} & \ranksecond{0.05}
& \ranksecond{0.16} & \ranksecond{0.08} \\

Drive-OccWorld~\cite{yang2025driveoccworld}
& C & O+F
& -- & --
& 0.25 & \rankthird{0.44}
& \rankthird{0.72} & \rankthird{0.47}
& \rankthird{0.03} & 0.08
& 0.22 & 0.11 \\

OccWorld-D~\cite{zheng2024occworld}
& C & O
& -- & 2.80
& 0.39 & 0.73 & 1.18 & 0.77
& 0.11 & 0.19 & 0.67 & 0.32 \\

World4Drive~\cite{zheng2025world4drive}
& C & None
& \ranksecond{70.18} & --
& \rankthird{0.23} & 0.47 & 0.81 & 0.50
& \ranksecond{0.02} & 0.12 & 0.33 & 0.16 \\

ResWorld~\cite{zhangResWorldTemporalResidual2026}
& C & None
& 80.77& \rankthird{10.3$^\ddagger$}& \rankfirst{0.17} & \rankfirst{0.32}
& \rankfirst{0.55} & \rankfirst{0.35}
& \rankfirst{0.01} & \rankfirst{0.02}
& \ranksecond{0.16} & \rankfirst{0.07} \\

\midrule

FusionAD~\cite{ye2023fusionad}
& C+L & M+B+Mo+Tr+O
& 141.60& 1.22$^\ddagger$& -- & -- & -- & 1.03
& 0.25 & 0.13 & 0.25 & 0.21 \\

SpaRC-AD~\cite{wolters2025sparcad}
& C+R & M+B+Mo+D
& -- & --
& 0.24 & 0.47 & 0.79 & 0.50
& \rankfirst{0.01} & \rankthird{0.06}
& 0.20 & \rankthird{0.09} \\

OccWorld-O~\cite{zheng2024occworld}
& GT-O & O
& \rankthird{72.39}& \rankfirst{18.0}& 0.32 & 0.61 & 0.98 & 0.64
& 0.06 & 0.21 & 0.47 & 0.24 \\

\midrule

\textbf{RiskWorld (ours)}
& C+M+GT-Tr & Risk/O+Coll.
& 90.81 & \ranksecond{11.5}& \ranksecond{0.18} & \ranksecond{0.37}
& \ranksecond{0.65} & \ranksecond{0.40}
& 0.06 & 0.08
& \rankfirst{0.14} & \rankthird{0.09} \\

\bottomrule
\end{tabular}
\endgroup

\smallskip
\begin{minipage}{\textwidth}
\footnotesize
Input refers to inference inputs. Aux. Sup. excludes
ego-trajectory supervision and generic backbone pretraining.
C: camera; L: LiDAR; R: radar; M: map; B: boxes; D: depth; Mo: motion;
Tr: tracks; O: occupancy; F: flow; Coll.: collision.
GT-Tr denotes current and historical ground-truth actor states.
FPS denotes inference throughput in frames per second on a single
NVIDIA RTX 4090 for nuScenes open-loop planning. Published values are
used where available. $^\ddagger$ marks local measurements of ResWorld,
and FusionAD using the authors' original implementations.
CR denotes collision rate. 
Parameter counts (Params.) refer to the online inference model where available.
Counts for VAD, DiffusionDrive, World4Drive, and ResWorld are computed
from released implementations or configurations. World4Drive excludes
models used only for offline depth and mask generation.
Gray shading marks the three best distinct values per metric. Bold
and underlining indicate first and second place, respectively, with
ties sharing the same formatting. Lower is better except for FPS.
Entries without comparable measurements (``--'') are excluded from
ranking. Input settings differ across methods.
\end{minipage}

\end{table*}

\begin{table}[!t]
\caption{Planning performance under inference-time perturbations
of RiskExpert features.}
\label{tab:risk_feature_ablation}
\centering
\footnotesize
\setlength{\tabcolsep}{4pt}
\renewcommand{\arraystretch}{1.15}

\begin{tabular*}{\columnwidth}{
@{\extracolsep{\fill}}l|r|r@{}}
\toprule
Risk features
& L2 Avg. (m) $\downarrow$
& Collision Avg. (\%) $\downarrow$ \\
\midrule
Original (Full)
& \textbf{0.40} & \textbf{0.09} \\
Zeroed
& 0.63 & 0.28 \\
Shuffled
& 0.49 & 0.27 \\
\bottomrule
\end{tabular*}
\end{table}
\begin{table}[!t]
\caption{Ablation of Persistence-referenced intervention (PRI)
and Flow Transport (FT).}
\label{tab:ablation}
\centering
\footnotesize
\setlength{\tabcolsep}{3pt}
\renewcommand{\arraystretch}{1.15}

\begin{tabular*}{\columnwidth}{
@{\extracolsep{\fill}}l|cc|r|rrrr@{}}
\toprule
Variant
& PRI
& FT
& L2 (m) $\downarrow$
& \multicolumn{4}{c}{CR (\%) $\downarrow$} \\
\cline{5-8}
& & & Avg.
& 1s & 2s & 3s & Avg. \\
\midrule
CSR
& $\times$ & --
& 0.40 & 0.08 & 0.10 & 0.17 & 0.12 \\

No-Transport
& $\checkmark$ & $\times$
& 0.40 & 0.08 & 0.10 & 0.18 & 0.12 \\
\midrule
\textbf{Full}
& $\checkmark$ & $\checkmark$
& 0.40 & \textbf{0.06} & \textbf{0.08}
& \textbf{0.14} & \textbf{0.09} \\
\bottomrule
\end{tabular*}
\par\vspace{1pt}
\smallskip
\begin{minipage}{\columnwidth}
\footnotesize
``--'' denotes not applicable.
Values are rounded for display. Relative reductions are computed
from the corresponding metric values before rounding.
\end{minipage}
\par\vspace{-9pt}
\end{table}

\subsection{Dataset and Metrics}
\label{sec:datasets_metrics}

We evaluate RiskWorld on the \textbf{nuScenes}
dataset~\cite{caesar2020nuscenes} for open-loop planning.
The evaluation subset contains 6,019 samples from
150 validation scenes, of which 5,119 have complete
future annotations and are used to compute planning
metrics. Performance is measured by trajectory L2 error (m)
and ego-box collision rate (\%), which quantify
deviation from the ground-truth trajectory and
collisions between the predicted ego footprint
and other actors, respectively.

Following the VAD/STP3
protocol~\cite{hu2022stp3,jiang2023vad},
results are reported at the horizons of 1, 2, and 3\,s.
Each horizon value averages the per-step measurements
up to that time; ``Avg.'' denotes the mean of
the three horizon values.

\subsection{Implementation Details}
\label{sec:implementation}

RiskWorld combines GeoBEV visual
features~\cite{zhang2025geobev}, extracted using a
ResNet-50 image backbone, with structured map and
actor information and ego history.
Planning uses a fixed set of 12 candidate trajectories,
each containing six future poses at 0.5\,s intervals
over a 3\,s horizon.

Training proceeds in stages. During final calibration, only the incoming-risk bridge is optimized
using AdamW with an initial learning rate of $10^{-4}$ and a global
batch size of 16. All other modules remain frozen.

\subsection{Main Results}
\label{sec:main_results}

\textbf{Planning performance:}
Table~\ref{tab:model_comparison} compares RiskWorld with
representative state-of-the-art methods on nuScenes.
RiskWorld achieves an average L2 error of 0.40\,m and
an average collision rate of 0.09\%.
Its average L2 error is lower than those of VAD~\cite{jiang2023vad},
SparseDrive~\cite{sunSparseDriveEndtoEndAutonomous2025}, and DiffusionDrive~\cite{liao2025diffusiondrive}.
At the 3\,s horizon, RiskWorld delivers the lowest
collision rate among the listed methods at 0.14\%,
compared with 0.16\% for both DiffusionDrive~\cite{liao2025diffusiondrive} and ResWorld~\cite{zhangResWorldTemporalResidual2026}.
These results demonstrate its competitive trajectory
accuracy and favorable collision performance at a long horizon evaluated.

\textbf{Efficiency comparison:}
Table~\ref{tab:model_comparison} includes the model size and inference
throughput for comparison. RiskWorld has 90.81\,M parameters, comparable to
SparseDrive-S~\cite{sunSparseDriveEndtoEndAutonomous2025} and
ResWorld~\cite{zhangResWorldTemporalResidual2026}, slightly fewer
than DiffusionDrive~\cite{liao2025diffusiondrive}, and fewer than
UniAD~\cite{hu2023uniad} and FusionAD~\cite{ye2023fusionad}.
Its reported throughput is 11.5\,FPS on a single NVIDIA RTX 4090,
the second-highest value listed and the highest among the listed
methods with camera inputs. OccWorld-O~\cite{zheng2024occworld}
reports 18.0\,FPS using ground-truth occupancy without image
processing, whereas its camera-based variant,
OccWorld-D~\cite{zheng2024occworld}, reports 2.8\,FPS.
Please note that throughput comparisons should account for differences in input
conditions and the processing stages included in each measurement.

\begin{figure*}[t]
    \centering
    \includegraphics[
        width=0.99\textwidth,
        trim=0 5 0 0,
        clip
    ]{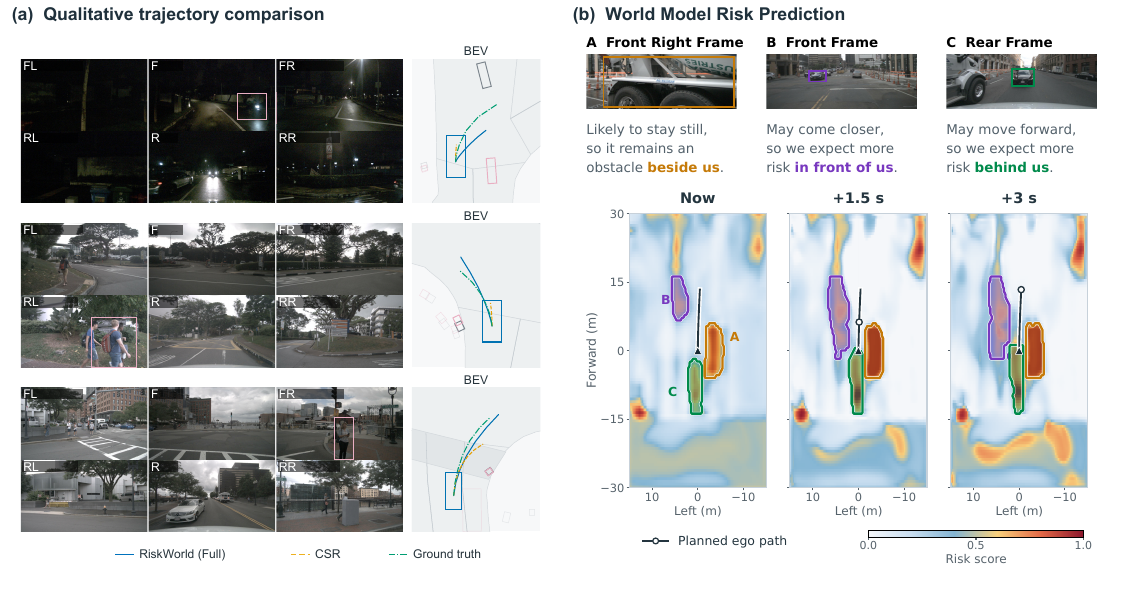}
    \vspace{-1.5mm}
    \caption{Qualitative comparison on nuScenes.
    (a) Six camera views and BEV trajectories from RiskWorld (blue solid),
    CSR (orange dashed), and ground truth (green dash-dotted). The blue box
    marks the initial ego footprint.
    (b) World Model's predicted collision occupancy evolution around the
    side (orange), front (purple), and rear (green) vehicles, with the
    planned ego path overlaid.}
    \vspace{-2mm}
    \label{fig:rw_qualitative}
\end{figure*}

\textbf{Qualitative comparison:}
Fig.~\ref{fig:rw_qualitative}(a) compares RiskWorld with CSR,
which relies on current-world evidence. Across a nighttime
road, a turn near pedestrians, and an urban intersection,
RiskWorld selects trajectories with greater forward progress
in the first two examples and follows the ground-truth turn
more closely in the third. Fig.~\ref{fig:rw_qualitative}(b) shows the predicted future occupancy of the surrounding scene. The construction truck to the right of the ego vehicle remains largely stationary, producing persistent risk in that region marked by an orange mask, while the vehicles from the opposing traffic flow and from behind the ego generate dynamically evolving risk over time. This distinction allows the model to capture both stationary constraints and moving hazards, providing future-risk evidence that can guide the direction and extent of trajectory adjustment.

\subsection{Ablation Studies}
\label{sec:ablation}
\textbf{RiskExpert features:}
Table~\ref{tab:risk_feature_ablation} examines how
planning depends on RiskExpert features.
We zero these features or shuffle them across
samples at inference, keeping model weights, visual inputs, candidate
trajectories, and the selector fixed. Zeroing and shuffling increase
average collision rates from 0.09\% to 0.28\% and 0.27\%, respectively, and average
L2 error from 0.40\,m to 0.63\,m and 0.49\,m.
These results indicate that planning depends on both the risk features and their correspondence
with the current scene.


\textbf{Future-risk evidence:}
We examine the contributions of future-risk evidence
and flow transport using the same input cache,
candidate set, evaluation samples, and model weights, changing only the
specified inference component.
Table~\ref{tab:ablation} compares the full RiskWorld model
(\textbf{Full}) with CSR, which selects the anchor using current-state
evidence without the additional future-risk correction.
\textbf{Full} achieves lower collision rates than CSR at all
three evaluation horizons, reducing the average
collision rate from 0.12\% to 0.09\%.
This comparison supports the use of future-risk evidence in candidate
selection.

\textbf{Flow transport:}
To assess the contribution of spatial transport, we evaluate
\emph{No-Transport}, a variant that sets the predicted flow to zero
at inference while retaining the same model weights and all other
evolution and selection components. Table~\ref{tab:ablation} reports a 3\,s collision rate of 0.18\% for
No-Transport, compared with 0.14\% for \textbf{Full}. \textbf{Full} reduces the average collision rate relative to
No-Transport, with essentially unchanged average L2 error. This fixed-checkpoint comparison supports the contribution of
spatial transport to future-risk assessment and trajectory selection.

\textbf{Temporal order:}
In a separate experiment using only current and past observations,
we fix the current frame and shuffle earlier frames. Chronological
history improves 3\,s incoming-cell Area Under
the Precision-Recall Curve (AUPRC) by 0.006 over shuffled
history (paired 95\% confidence interval: [0.005, 0.008]).
This supports the value of temporal order for predicting incoming
occupancy, which informs downstream risk assessment.

\textbf{Candidate-count scalability:}
With BEV features and candidate trajectories provided,
increasing the candidate count from $K=1$ to $K=24$
changes the shared world prediction time from
13.76\,ms to 13.87\,ms and the total runtime from
44.16\,ms to 45.20\,ms.
These measurements exclude the visual frontend.
The small increase in runtime supports reusing a
shared world forecast to evaluate additional candidates
at low marginal cost.

\section{CONCLUSIONS}

In this paper, we introduce RiskWorld, a risk-aware world modeling framework for
shared occupancy forecasting and selective trajectory replacement.
By integrating spatial risk fields and temporal actor context with
visual BEV features, RiskWorld combines flow-guided scene evolution
with trajectory-level risk assessment. Comparing future evidence
with a current-state persistence reference guides changes to the
planning anchor under explicit constraints on predicted risk and
trajectory error, while preserving candidate geometry. Experiments
on nuScenes show competitive trajectory accuracy and the lowest
reported collision rate at the longest evaluated horizon (3\,s)
among among all evaluated state-of-the-art baselines.
Within-setting comparisons further demonstrate reduced collision rates relative to current-state selection,
while sharing a single forecast enables additional candidates to be
evaluated at low marginal computational cost. These results support
the use of future-risk evidence for efficient, selective planning
intervention. Future work should address candidate-dependent actor
responses and prediction uncertainty in reactive closed-loop planning
while preserving efficient forecast reuse.



\bibliographystyle{IEEEtran}
\bibliography{references}
\end{document}